\documentclass[letterpaper]{article} % DO NOT CHANGE THIS
\usepackage[preprint]{aaai2027}  % DO NOT CHANGE THIS
\usepackage[hyphens]{url}  % DO NOT CHANGE THIS
\usepackage{graphicx} % DO NOT CHANGE THIS
\usepackage{enumitem}
\usepackage{natbib}  % DO NOT CHANGE THIS AND DO NOT ADD ANY OPTIONS TO IT
\usepackage{caption} % DO NOT CHANGE THIS AND DO NOT ADD ANY OPTIONS TO IT
\usepackage{algorithm}
\usepackage{algpseudocode}
\usepackage{amsmath}
\usepackage{amssymb}
\usepackage{newfloat}
\usepackage{listings}
\DeclareCaptionStyle{ruled}{labelfont=normalfont,labelsep=colon,strut=off} % DO NOT CHANGE THIS
\floatstyle{ruled}
\newfloat{listing}{tb}{lst}{}
\floatname{listing}{Listing}

\usepackage{booktabs}

\title{Stop When Memory Suffices: Evidence-Conditioned Progressive Execution \\ for LLM Agents}
\author{
    Yidan Lin,
    Kaixiang Wang,
    Jiong Lou,
    Jie Li
}
\affiliations{
}

\begin{document}

\maketitle

\begin{abstract}
The continued development of LLMs toward persistent and adaptive intelligence increasingly requires long-term memory mechanisms that preserve and reuse information across interactions. Existing memory systems either compress and structure histories for efficient access or perform deep research over broader trajectories. The former lowers online cost but may omit temporal, causal, or cross-step dependencies, while the latter improves evidence coverage at substantial latency and inference cost. This raises a key question: can a memory system achieve strong answer quality while maintaining low online latency? We introduce Router-Mem, an evidence-conditioned progressive execution framework for long-horizon agent memory. Router-Mem first applies a shared low-cost retrieval prefix to obtain evidence. A lightweight sufficiency router then predicts whether the context supports early termination, which enable a single-token decision at inference time. It is trained with evidence-level supervision and rationale-conditioned representation distillation. When evidence is insufficient, Router-Mem reuses retrieval hits to expand memory blocks and perform deeper analysis and aggregation. Experiments on AMA-Bench and BEAM show that Router-Mem achieves 55.17\% and 38.77\% score while reducing average inference time by 27.3\% and 25.5\% compared with full memory execution.
\end{abstract}

\section{Introduction}

Large language models are evolving from static generators into interactive systems that reason, use tools, learn from feedback, and operate over extended tasks~\cite{yao2023react,schick2023toolformer,shinn2023reflexion}. This creates a growing need to retain and reuse information beyond a single context window~\cite{wang2024voyager,zhou2024webarena}. Long-running applications accumulate conversations, tool calls, observations, errors, and intermediate decisions~\cite{liu2024agentbench,yang2024sweagent}. Later queries must recover relevant evidence while preserving temporal and cross-step dependencies~\cite{li2024survey,guo2024large}. Although recurrent memory, sparse attention, and efficient exact attention have extended context capacity~\cite{dai2019transformerxl,beltagy2020longformer}, full-history processing remains costly~\cite{zaheer2020bigbird,dao2022flashattention}, and performance is still sensitive to evidence position, sequence length, and task complexity~\cite{liu2024lost,bai2024longbench}. These limitations motivate dedicated memory mechanisms for selectively retrieving and organizing query-relevant experience~\cite{hsieh2024ruler,zhang2024infinitebench}.

\begin{figure}[t]
    \centering
    \includegraphics[width=\linewidth]{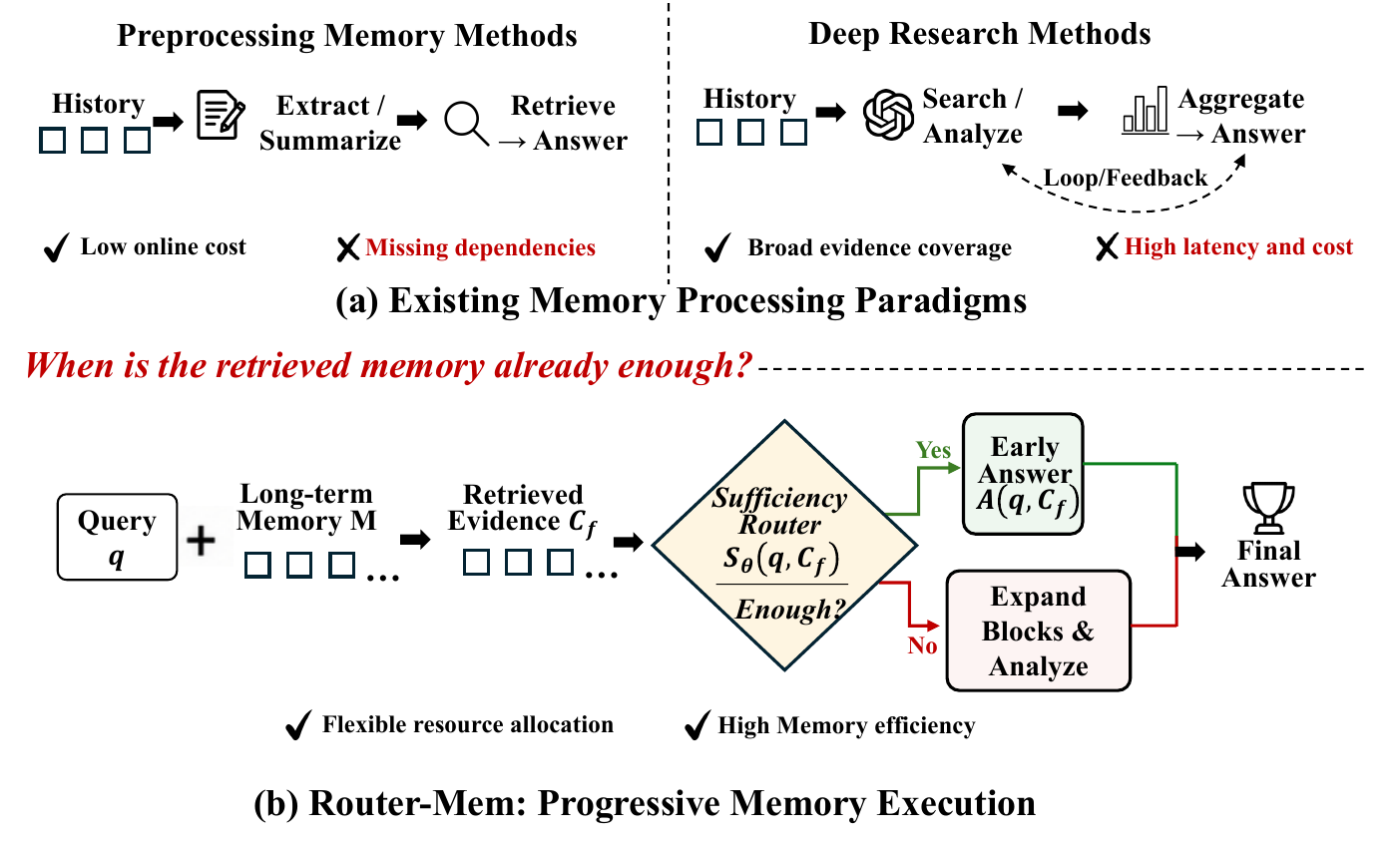}
    \caption{The Existing Memory Paradigms VS Router-mem.}
    \label{fig:audit_dilemma}
\end{figure}

Existing agent memory systems broadly follow two processing paradigms. Preprocessing-oriented methods extract, summarize, or restructure interaction histories into compact representations for efficient access~\cite{zhang2025survey,hu2025memory}(as shown in Fig.~\ref{fig:audit_dilemma}a). Mem0~\cite{chhikara2025mem0} consolidates salient conversational facts, A-MEM~\cite{xu2025amem} organizes memories as dynamically linked notes, and G-Memory~\cite{zhang2025gmemory} builds hierarchical graphs for multi-agent histories. These designs reduce online retrieval cost, but early compression may omit temporal, causal, or cross-step details needed by future queries. Query-time reconstruction methods instead retain broader histories and perform iterative search and reasoning after a query arrives. GAM~\cite{yan2025gam} uses a researcher agent to construct query-specific context, while E-mem~\cite{wang2026ememmultiagentbasedepisodic} activates episodic segments and aggregates analyses from multiple memory agents. They improve evidence coverage, but can impose unnecessary latency when local memory already suffices, exposing a fundamental quality--efficiency trade-off~\cite{du2026memory}.

% These paradigms need not remain fixed. The required depth of memory processing depends on the evidence recovered for the current query, rather than on the query alone. Adaptive RAG methods route by predicted complexity or decide whether to continue retrieval from intermediate signals~\cite{jeong2024adaptive,park2025stoprag,li2026raser}. However, they mainly target document question answering and do not study progressive execution over heterogeneous agent histories. This raises a central question: can a shared retrieval prefix determine when broader memory reconstruction is no longer necessary?

However, memory-query processing need not follow a fixed paradigm. Queries whose supporting evidence is already localized should not incur the cost of deep memory research, whereas queries with incomplete or
dispersed evidence may not be answered reliably through compact retrieval alone. This raises a central question: can a memory system infer the required processing depth from the query and the evidence recovered by lightweight retrieval, thereby allocating computation more effectively?

To address this gap, we propose Router-Mem, an evidence-conditioned progressive execution framework for long-horizon agent memory (as shown in Fig.~\ref{fig:audit_dilemma}b). For every query, Router-Mem first runs a shared low-cost retrieval prefix to obtain an initial evidence context. This prefix is shared by both execution paths and provides the key evidence for deciding whether the system can return early. A lightweight sufficiency router then evaluates the query together with the retrieved context and predicts whether the current evidence supports early termination. The router is trained with evidence-level supervision and rationale distillation, which transfers complex sufficiency judgments into a single-token decision at inference time. When the retrieved evidence is sufficient, the system answers directly and skips the remaining computation. Otherwise, it reuses the retrieval hits as localization signals, expands the relevant memory blocks, and invokes deeper analysis and evidence aggregation. The fast and slow paths are coupled through the shared prefix, which avoids restarting memory search from scratch. This design preserves inexpensive access for locally answerable queries while allocating costly memory reasoning to cases that still lack decisive evidence, yielding a better balance between answer quality, latency, and inference cost.

Experiments on AMA~\cite{zhao2026amabench} and BEAM~\cite{tavakoli2026beam} show that Router-Mem achieves 55.17\% and 38.77\% score, while reducing average inference time by 27.3\% and 25.5\% relative to full memory processing. Across different routing thresholds, it consistently provides favorable quality--latency trade-offs and forms a strong empirical Pareto frontier among compared methods.

Our main contributions are as follows:
\begin{itemize}[leftmargin=2em]
    \item We formulate long agentic memory based query as an evidence-conditioned progressive execution problem. The required processing depth is determined by the query and the evidence recovered at runtime.

    \item We propose Router-Mem, an evidence-conditioned progressive framework that addresses fixed computation in long-horizon memory access. It combines a reusable retrieval prefix, evidence-sufficiency routing, and retrieval-guided continuation to stop when evidence is sufficient and invoke broader memory analysis only when needed, balancing efficiency with deep memory reasoning quality.

    \item We evaluate Router-Mem on AMA-Bench and BEAM across different routing thresholds and model settings. The results demonstrate strong answer quality with substantially lower inference time and establish a favorable quality--latency Pareto frontier.
\end{itemize}

\section{Related Work}

\paragraph{Retrieval-Augmented Generation.}
Retrieval-augmented generation (RAG) grounds LLMs in external knowledge and has become a standard paradigm for knowledge-intensive tasks~\cite{lewis2020rag}. Some works like Adaptive-RAG, Stop-RAG, further selects among no retrieval, single-step retrieval, and iterative retrieval according to predicted query complexity~\cite{jeong2024adaptive,park2025stoprag,li2026raser}. However, this query-level routing does not account for the evidence actually recovered at runtime: a difficult query may be resolved by successful initial retrieval, whereas a seemingly simple query may still lack decisive evidence. Recent work also improves retrieval through structured representations. HippoRAG combines knowledge graphs with Personalized PageRank for associative and multi-hop retrieval~\cite{gutierrez2024hipporag}, while GraphRAG and LightRAG use graph-based indexing for global and hierarchical knowledge discovery~\cite{edge2024graphrag,guo2024lightrag}. Despite improving retrieval planning and evidence organization, these methods do not directly determine whether the currently retrieved evidence is sufficient to terminate further computation.

% Retrieval-augmented generation (RAG) grounds LLMs on external knowledge and has become a standard paradigm for knowledge-intensive tasks~\cite{lewis2020rag}. Beyond conventional retrieve-then-generate pipelines, adaptive RAG methods dynamically adjust retrieval strategies according to query complexity or intermediate signals~\cite{jeong2024adaptive}. Recent studies further enhance retrieval with structured representations: HippoRAG integrates knowledge graphs with Personalized PageRank for associative and multi-hop retrieval~\cite{gutierrez2024hipporag}, while GraphRAG and LightRAG leverage graph-based indexing for global and hierarchical knowledge discovery~\cite{edge2024graphrag,guo2024lightrag}. Despite improved evidence organization and retrieval, these approaches mainly optimize how evidence is retrieved, rather than determining whether the currently retrieved evidence is already sufficient to terminate further computation.

\paragraph{Memory Systems for LLM Agents.}
Long-term agent memory systems aim to efficiently preserve and access past experiences. Preprocessing-based approaches construct compact memory representations before queries arrive. Mem0 extracts and consolidates salient conversational information~\cite{chhikara2025mem0}, A-MEM organizes experiences as evolving structured notes~\cite{xu2025amem}, and G-Memory builds hierarchical graphs for multi-agent interactions~\cite{zhang2025gmemory}. ReasoningBank further distills reusable reasoning strategies from agent trajectories~\cite{ouyang2026reasoningbank}. These methods reduce retrieval cost but may lose fine-grained temporal and causal dependencies. In contrast, query-time reconstruction methods retain broader histories and perform deeper reasoning when queries arrive. GAM employs researcher agents to construct query-specific contexts~\cite{yan2025gam}, while E-mem uses episodic retrieval and multi-agent analysis for context reconstruction~\cite{wang2026ememmultiagentbasedepisodic}. Although these methods improve evidence coverage, they may introduce unnecessary computation when lightweight retrieval already provides sufficient evidence.

Our work connects these two regimes through a shared retrieval prefix and an evidence-sufficiency router, which terminates early when the current context supports an answer and preserves deeper processing for queries that require it.

\begin{figure}[t]
    \centering
    \includegraphics[width=0.9\linewidth]{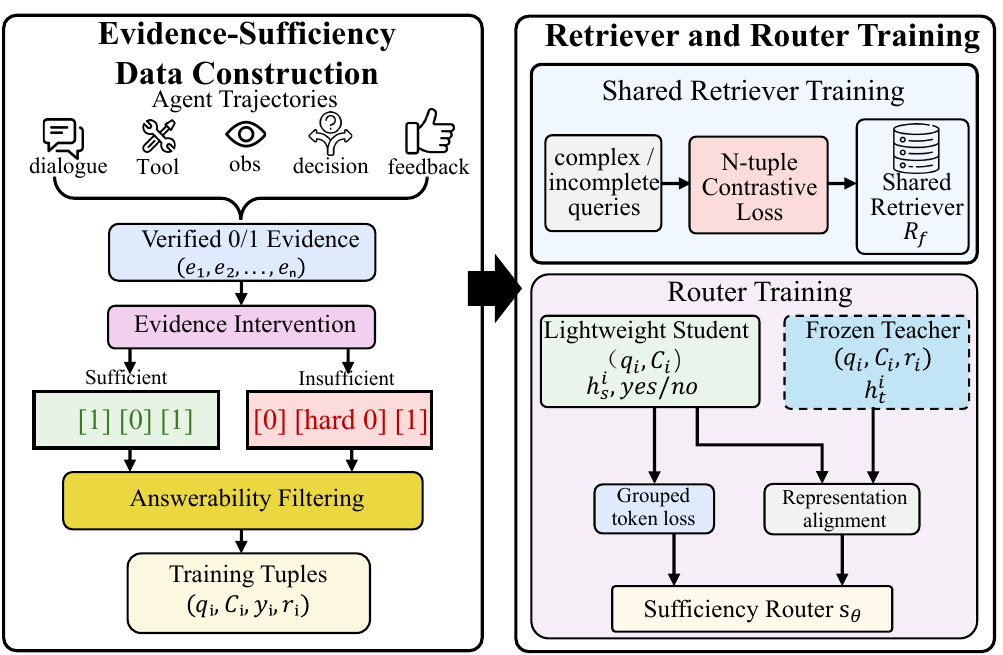}
    \caption{Training Process of Router-mem.}
    \label{fig:train}
\end{figure}

\section{Method}

In this section we systematically presents the design of Router-Mem including training process and system pipeline. More details can be found in \textbf{technical supplement}. 

\subsection{Progressive Memory Execution}
\label{sec:progressive_execution}

Long-horizon agents continuously accumulate heterogeneous memories, including conversations, intermediate decisions, tool calls, environment observations, and execution feedback. Queries over such memories require different levels of processing. Some queries can be answered from a small set of locally relevant memories recovered by lightweight retrieval. Others depend on evidence that is dispersed across time, execution steps, or interaction trajectories, and therefore require broader memory expansion and aggregation. Applying only lightweight retrieval to every query may miss such dependencies, whereas executing the full memory-processing pipeline for every query introduces unnecessary latency and inference cost.

We therefore organize memory access as a progressive execution process. Every query first performs a shared low-cost retrieval prefix. If the retrieved context already contains sufficient evidence, the system directly generates an answer and skips the remaining computation. Otherwise, it continues with broader memory processing. The fast and slow paths are coupled through the shared retrieval prefix, as the expensive stage continues from the retrieved evidence and reuses its localization signals for broader memory processing.

Determining whether the retrieved context is sufficient is itself non-trivial. A direct solution is to prompt a general-purpose LLM to inspect the query and retrieved memories and explicitly reason about whether further processing is needed. However, a reliable judgment may require checking for missing evidence, temporal updates, action--feedback relations, and conflicting records. Although the final output is only a binary decision, the underlying reasoning can be complex. Performing such deliberation online often requires a capable model, a long context prefill, and additional reasoning tokens. This cost is undesirable because the router is introduced precisely to avoid unnecessary computation. An expensive routing decision may substantially reduce the efficiency gain obtained from early termination.

To address this issue, we train a lightweight evidence-sufficiency router that directly predicts whether the current retrieval evidence state supports early termination. Given a query $q$ and an ordered memory
$M=(z_1,\ldots,z_T)$, the shared retriever first returns

\begin{equation}
C_f = R_f(q,M),
\end{equation}

where $C_f$ is the initially retrieved memory context. We define the context-dependent termination label as

\begin{equation}
y(q,C_f)
=
\mathbf{1}
\left[
C_f \text{ provides sufficient evidence to answer } q
\right],
\end{equation}

and train the router to estimate

\begin{equation}
s_{\theta}(q,C_f)
=
P_{\theta}
\left(
y=1 \mid q,C_f
\right).
\end{equation}

Importantly, the decision is conditioned on the execution state $(q,C_f)$ rather than on the query alone. The same query may terminate after one retrieval result but require further processing after another result that omits a decisive memory step. During training, we use evidence-level supervision and rationale distillation to transfer the required evidence judgment into the lightweight router. At inference time, the router reads only $(q,C_f)$ and produces a single-token termination decision.

\subsection{Evidence-Sufficiency Data Construction}
\label{sec:data_construction}

Training the router requires supervision that distinguishes evidence sufficiency from semantic relevance (as shown in fig.~\ref{fig:train}). We collect long-horizon agent trajectories from diverse interaction settings, including tool use, software engineering, web navigation, games, and long conversations. These training sources are disjoint from the downstream evaluation benchmarks, while preserving similar interaction patterns such as multi-step actions, environment feedback, and temporally dispersed evidence.

For each trajectory, we construct and verify a query--answer--evidence tuple $(q_i,a_i,E_i)$, where $E_i$ contains the memory steps that support the reference answer $a_i$. We then apply evidence intervention to produce contexts with different sufficiency labels. A positive context retains all required evidence and mixes it with distractor steps:
\begin{equation}
C_i^{+}=E_i\cup D_i,
\end{equation}
where $D_i$ contains randomly sampled memory steps from the same trajectory. Since the complete supporting evidence is preserved, $C_i^{+}$ is labeled as sufficient.

A negative context removes part or all of the required evidence and fills the remaining context with semantically related hard distractors:
\begin{equation}
C_i^{-}
=
\left(E_i\setminus E_i^{\mathrm{drop}}\right)
\cup D_i^{\mathrm{hard}}.
\end{equation}
The hard distractors may share entities, tools, or nearby events with the query, but do not preserve the complete evidence chain. These examples teach the router that relevance alone does not imply answerability.

Evidence removal does not always make a context insufficient, because redundant or alternative support may remain. We therefore evaluate each candidate negative with multiple answer models and discard it if any model still produces a correct answer. For every retained sample, a teacher model generates a rationale that explains which evidence is present and which decisive information is missing. The resulting dataset consists of tuples $(q_i,C_i,y_i,r_i)$, where $y_i\in\{\mathrm{yes},\mathrm{no}\}$ is the sufficiency label and $r_i$ is the teacher rationale. We split the data by trajectory to prevent related questions, neighboring steps, and context variants from leaking across training and validation sets.

\subsection{Training Process}
\label{sec:router_training}

Given a constructed sample $(q_i,C_i,y_i,r_i)$, we form the student input as
\begin{equation}
x_i=\operatorname{Prompt}(q_i,C_i).
\end{equation}
The router retains the causal language-model interface and predicts the sufficiency label through its next-token distribution. Let $V_{\mathrm{yes}}$ and $V_{\mathrm{no}}$ denote token groups corresponding to the normalized outputs ``yes'' and ``no.'' We optimize the grouped-token classification loss
\begin{equation}
\mathcal{L}_{\mathrm{cls}}^{i}
=
-
\log
\sum_{v\in V_{y_i}}
P_{\theta}(v\mid x_i).
\end{equation}

The binary label specifies the desired decision but provides limited supervision about the underlying evidence judgment. We therefore use rationale-conditioned representation distillation. A frozen teacher additionally reads the rationale $r_i$, while the student observes only $(q_i,C_i)$. Let $h_s^i$ denote the student representation before the decision token, and let $h_t^i$ denote the teacher representation after processing the rationale. We align them at a selected intermediate layer using
\begin{equation}
\mathcal{L}_{\mathrm{rep}}^{i}
=
1-
\frac{
(h_s^i)^{\top}h_t^i
}{
\lVert h_s^i\rVert_2
\lVert h_t^i\rVert_2
}.
\end{equation}

The final training objective is
\begin{equation}
\mathcal{L}_{\mathrm{router}}
=
\mathcal{L}_{\mathrm{cls}}
+
\lambda_{\mathrm{rep}}s_0
\mathcal{L}_{\mathrm{rep}},
\end{equation}
where $\lambda_{\mathrm{rep}}$ controls the contribution of representation distillation and $s_0$ calibrates the scales of the two losses. The teacher and rationale are used only during training. At inference time, the student reads $(q,C_f)$ and produces a single decision token without generating an explanation.

\begin{figure}[t]
    \centering
    \includegraphics[width=0.9\linewidth]{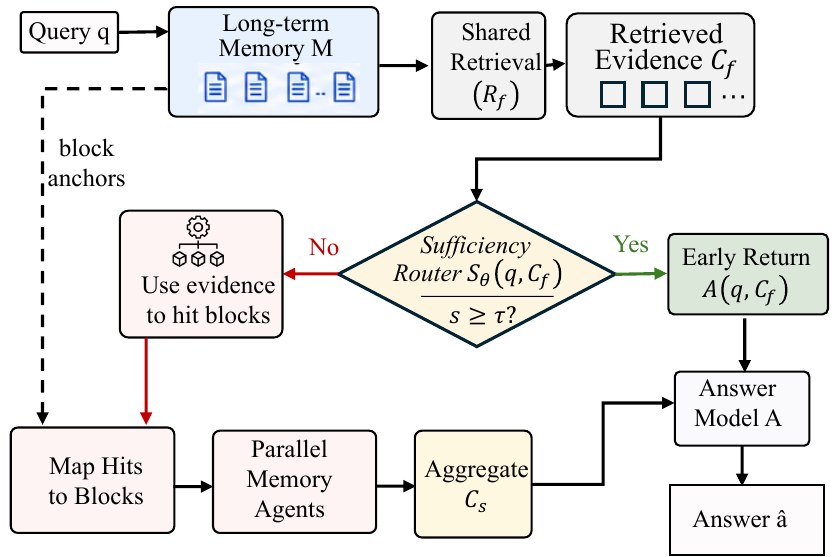}
    \caption{System Pipeline of Router-mem.}
    \label{fig:system}
\end{figure}

\begin{table*}[t]
\centering
\caption{Results on AMA-Bench with two backbone LLMs: DeepSeek-V4-Flash and Qwen3.5-35B-A3B.}
\label{tab:ama_main}
\scriptsize
\setlength{\tabcolsep}{2pt}
\begin{tabular}{lccccc|ccccc}
\toprule
& \multicolumn{5}{c|}{\textsc{DeepSeek-V4-Flash}} &
  \multicolumn{5}{c}{\textsc{Qwen3.5-35B-A3B}} \\
\cmidrule(lr){2-6}\cmidrule(l){7-11}
Method & Score (\%) & Early (\%) & $T_{\rm mem}$ (s) & $T_{\rm e2e}$ (s) & $T_{\rm build}$ (s)
       & Score (\%) & Early (\%) & $T_{\rm mem}$ (s) & $T_{\rm e2e}$ (s) & $T_{\rm build}$ (s) \\
\midrule
\multicolumn{11}{l}{\textit{External Memory Baselines}} \\
RAG          & 40.46 & -- & 0.057 & 2.49  & 10.15  & 45.21 & -- & 0.059 & 4.72 & 9.88 \\
LightMem     & 35.34 & -- & 0.058           & 2.30  & 1157.78  & 36.58 & -- & 0.055 & 2.49 & 6531.13 \\
Mem0         & 35.90 & -- & 0.159           & 3.50  & 46.12  & 43.96 & -- & 0.156 & 2.61 & 51.37 \\
Long Context & 47.08 & -- & 3.595           & 5.18  & 0.001 & 53.95 & -- & 20.520 & 22.80 & 0.002 \\
GAM          & 40.21 & -- & 22.450          & 25.289& 123.57 & 43.96 & -- & 24.497 & 28.40 & 141.52 \\
\midrule
\multicolumn{11}{l}{\textit{Fixed Execution Endpoints}} \\
Fast Prefix Only        & 44.91 & 100.0 & 0.054 & 3.70  & 5.07 & 48.09 & 100.0 & 0.054 & 3.63 & 4.56 \\
Full Completion (Slow) & 56.49 & 0.0 & 11.341 & 14.47 & 5.07 & 56.41 & 0.0 & 15.816 & 20.28 & 4.66 \\
\midrule
\multicolumn{11}{l}{\textit{Router-Mem: Evidence-Conditioned Early Termination}} \\
Router-Mem ($\tau=0.1$) & 51.48 & 53.7 & 5.608  & 9.21  & 4.93 & 52.50 & 54.6 & 8.414  & 12.80 & 4.57 \\
Router-Mem ($\tau=0.3$) & 52.68 & 38.5 & 7.215  & 10.53 & 4.90 & 53.65 & 39.2 & 9.351  & 13.10 & 4.65 \\
Router-Mem ($\tau=0.5$) & 55.17 & 30.0 & 8.248  & 11.57 & 5.08 & 55.06 & 30.2 & 10.581 & 14.49 & 4.60 \\
Router-Mem ($\tau=0.7$) & 54.73 & 22.1 & 9.045  & 12.26 & 4.98 & 55.52 & 22.3 & 13.888 & 18.67 & 4.71 \\
Router-Mem ($\tau=0.9$) & 55.89 & 12.3 & 10.273 & 13.51 & 4.94 & 55.52 & 12.5 & 14.423 & 18.96 & 4.61 \\
\bottomrule
\end{tabular}
\end{table*}
\paragraph{Shared retriever training.}
We train the shared retriever with the same QA--evidence supervision. Evidence-complete memory views are treated as positives, while semantically related but incomplete views are used as negatives. Using an L2-normalized encoder $e(\cdot)$, we optimize a multi-positive contrastive objective:
\begin{equation}
\mathcal{L}_{\mathrm{ret}}^{i}
=
-
\frac{1}{|P_i|}
\sum_{p\in P_i}
\log
\frac{
\exp\left(e(q_i)^{\top}e(p)/T\right)
}{
\sum_{c\in\mathcal{C}_i}
\exp\left(e(q_i)^{\top}e(c)/T\right)
}.
\end{equation}
This objective encourages the shared prefix to retrieve answer-supporting evidence. 

\begin{algorithm}[t]
\caption{Progressive Memory Execution}
\label{alg:progressive_memory}
\begin{algorithmic}[1]
\Require Query $q$, memory $M$, threshold $\tau$
\Ensure Answer $\hat{a}$

\State $(C_f,H_f) \gets R_f(q,M)$
\Comment{Shared retrieval prefix}

\If{$C_f$ fits the router input budget}
    \State $s \gets s_{\theta}(q,C_f)$
    \Comment{Evidence-sufficiency score}
    \If{$s \geq \tau$}
        \State $\hat{a} \gets A(q,\operatorname{Order}(C_f))$
        \State \Return $\hat{a}$
        \Comment{Optimistic early return}
    \EndIf
\EndIf

\State $\mathcal{B}_f \gets \operatorname{MapToBlocks}(H_f)$
\State $\mathcal{B}_s \gets
       \operatorname{SelectBlocks}(\mathcal{B}_f,M)$

\ForAll{$b \in \mathcal{B}_s$ \textbf{in parallel}}
    \State $u_b \gets \operatorname{MemAgent}(q,b)$
\EndFor

\State $C_s \gets
       \operatorname{Aggregate}
       \left(\{u_b : b \in \mathcal{B}_s\}\right)$
\State $\hat{a} \gets A(q,C_s)$
\State \Return $\hat{a}$

\end{algorithmic}
\end{algorithm}

\subsection{Progressive Memory System}
\label{sec:system}

After training the retriever and termination router, we integrate them into a progressive memory execution system. The system first constructs aligned retrieval and processing views of the same memory (as shown in fig.~\ref{fig:system}). It then performs a shared retrieval step for every query and invokes broader memory processing only when the retrieved evidence is predicted to be insufficient.

\paragraph{Aligned memory views.}
Given an ordered memory $M=(z_1,\ldots,z_T)$, we construct two aligned views. The retrieval view indexes individual memory steps and stores their corresponding step and block identifiers. The processing view groups temporally adjacent steps into overlapping memory blocks. This alignment allows each retrieval hit to directly locate the memory blocks used by subsequent processing. The two views therefore support different stages of one execution process, rather than two independent memory systems.

\paragraph{Shared retrieval and termination.}
For a query $q$, the shared retriever first returns an initial context $C_f$ and its retrieval hits $H_f$:
\begin{equation}
C_f,H_f = R_f(q,M).
\end{equation}
The router evaluates the sufficiency of $C_f$ through its next-token distribution. We aggregate the probability assigned to the ``yes'' and ``no'' token groups:
\begin{equation}
p_{\mathrm{yes}}
=
\sum_{v\in V_{\mathrm{yes}}}
P_{\theta}(v\mid q,C_f),
\qquad
p_{\mathrm{no}}
=
\sum_{v\in V_{\mathrm{no}}}
P_{\theta}(v\mid q,C_f),
\end{equation}
and compute the normalized termination score
\begin{equation}
s(q,C_f)
=
\frac{p_{\mathrm{yes}}}
{p_{\mathrm{yes}}+p_{\mathrm{no}}}.
\end{equation}
Given a threshold $\tau$, the system follows
\begin{equation}
\pi_{\tau}(q,C_f)
=
\begin{cases}
\mathrm{terminate}, & s(q,C_f)\geq\tau,\\
\mathrm{continue}, & s(q,C_f)<\tau.
\end{cases}
\end{equation}
The threshold is selected on development data and fixed during evaluation. Varying $\tau$ yields different empirical trade-offs between answer quality and execution cost. If the retrieved context exceeds the router input budget, the system conservatively selects continuation.

\paragraph{Early return and execution continuation.}
If the termination condition is satisfied, the retrieved steps are restored to their original temporal order and passed to the answer model:
\begin{equation}
\hat{a}_f=A(q,C_f).
\end{equation}
This branch is an early return from the full memory execution, rather than the output of an independently selected fast system.

Otherwise, the system reuses the retrieval hits to continue the same execution process. Let $\mathcal{B}_f$ denote the blocks associated with $H_f$. We expand these anchors into a bounded set of processing blocks:
\begin{equation}
\mathcal{B}_s
=
\operatorname{SelectBlocks}(\mathcal{B}_f,M).
\end{equation}
The selected blocks include the retrieved anchor regions and their temporal neighborhoods, which helps recover local action--feedback and cross-step dependencies. Each block is analyzed independently by a memory agent:
\begin{equation}
u_b=\operatorname{MemAgent}(q,b),
\qquad b\in\mathcal{B}_s.
\end{equation}
An aggregator combines the block-level results into a consolidated context:
\begin{equation}
C_s
=
\operatorname{Aggregate}
\left(
\{u_b:b\in\mathcal{B}_s\}
\right),
\qquad
\hat{a}_s=A(q,C_s).
\end{equation}
The continuation stage therefore preserves and extends the computation performed by the shared retrieval prefix. It is invoked only when the current evidence does not support early termination.

\begin{table*}[t]
\centering
\caption{
End-to-end results on BEAM with Model DeepSeek-V4-Flash.
}
\label{tab:beam-main}
\small
\setlength{\tabcolsep}{6pt}
\begin{tabular}{lccccc}
\toprule
Method
& Score (\%) 
& Early Term. (\%) 
& $T_{\mathrm{mem}}$ (s) 
& $T_{\mathrm{e2e}}$ (s) 
& $T_{\mathrm{build}}$ (s) 
\\
\midrule

\multicolumn{6}{l}{\textit{External Memory Baselines}}\\

RAG
& 30.29
& --
& 0.094
& 1.66
& 217.01
\\

LightMem
& 27.12
& --
& 0.058
& 1.55
& 1174.31
\\

Mem0
& 22.09
& --
& 0.151
& 1.75
& 982.20
\\

Long Context
& 38.01
& --
& 14.304
& 17.08
& 0.05
\\
GAM & 35.23 & -- & 25.817 & 28.435 & 1334.78 \\
\midrule

\multicolumn{6}{l}{\textit{Fixed Execution Endpoints}}\\

Fast Prefix Only
& 30.02
& 100.0
& 0.054
& 2.51
& 31.81
\\

Full Completion (Slow)
& 43.26
& 0.0
& 14.611
& 18.43
& 31.90
\\

\midrule

\multicolumn{6}{l}{\textit{Router-Mem: Evidence-Conditioned Early Termination}}\\

Router-Mem ($\tau=0.1$)
& 33.94
& 56.9
& 5.675
& 8.51
& 32.21
\\

Router-Mem ($\tau=0.3$)
& 36.53
& 39.4
& 8.781
& 11.72
& 31.84
\\

Router-Mem ($\tau=0.5$)
& 38.77
& 26.6
& 10.883
& 14.11
& 31.88
\\

Router-Mem ($\tau=0.7$)
& 39.81
& 16.7
& 12.512
& 16.18
& 32.01
\\

Router-Mem ($\tau=0.9$)
& 42.09
& 8.0
& 13.422
& 16.93
& 31.86
\\

\bottomrule
\end{tabular}
\end{table*}

\section{Experiment}
We evaluate Router-Mem against state-of-the-art memory systems for LLM agents, focusing on effectiveness, routing efficiency, and robustness. Our experiments investigate three key research questions:

\textbf{RQ1: Comparative Effectiveness.}
Does Router-Mem provide advantages over existing memory mechanisms in terms of answer quality, inference latency, and overall efficiency?

\textbf{RQ2: Quality--Latency Trade-off.}
Can Router-Mem achieve a favorable balance between memory answer performance and inference latency, and how does the routing threshold control this trade-off?

\textbf{RQ3: Training Effectiveness.}
Do the proposed router training and embedding training improve retrieval quality, routing decisions, and end-to-end system performance?

\subsection{Experimental Setup.}

\paragraph{Dataset.}
We evaluate Router-Mem on two representative long-term memory benchmarks for LLM agents: AMA-Bench~\cite{zhao2026amabench} and BEAM~\cite{tavakoli2026beam}. AMA-Bench is designed to evaluate memory capabilities in long-horizon agentic scenarios, covering diverse interactions where agents must retrieve and reason over previously observed experiences. It emphasizes accurate memory access under complex multi-step dependencies. BEAM focuses on extremely long-context memory understanding and evaluates whether agents can effectively recover relevant information from large-scale conversational histories. Together, these benchmarks cover complementary challenges: AMA-Bench tests structured agent memory reasoning, while BEAM stresses scalability under massive context lengths. They provide a comprehensive evaluation of both memory effectiveness and efficient execution.

\paragraph{Settings.} We compare Router-Mem with representative memory baselines, including standard RAG~\cite{lewis2020rag}, compact memory systems such as LightMem~\cite{fang2026lightmemlightweightefficientmemoryaugmented} and Mem0~\cite{chhikara2025mem0}, Long Context, and deep query-time memory reconstruction methods GAM~\cite{yan2025gam}. For evaluation, we use DeepSeek-V4-Flash as the backbone model and report answer quality measured by the official LLM judge provided by each benchmark. We evaluate both efficiency and effectiveness by reporting memory construction time, retrieval latency, and end-to-end inference latency. Further experiments analyze the impact of routing thresholds $\tau$, router training components, and embedding model choices to evaluate the robustness and scalability of Router-Mem. We also evaluate the E2E token cost of Router-mem against existing agentic memory baselines in \textbf{technical supplement}.

% \begin{figure*}[t]
%     \centering

%     \begin{minipage}[t]{0.48\textwidth}
%         \centering
%         \includegraphics[width=0.85\linewidth]{Figures/fig2a_ama_pareto.pdf}
%         \centerline{(a) AMA-Bench}
%     \end{minipage}
%     \hfill
%     \begin{minipage}[t]{0.48\textwidth}
%         \centering
%         \includegraphics[width=0.85\linewidth]{Figures/fig2b_beam_pareto.pdf}
%         \centerline{(b) BEAM}
%     \end{minipage}

%     \caption{
%     Quality--latency trade-offs under different routing thresholds.
%     }
%     \label{fig:pareto}
% \end{figure*}

\begin{figure}[t]
    \centering
    \includegraphics[width=0.9\linewidth]{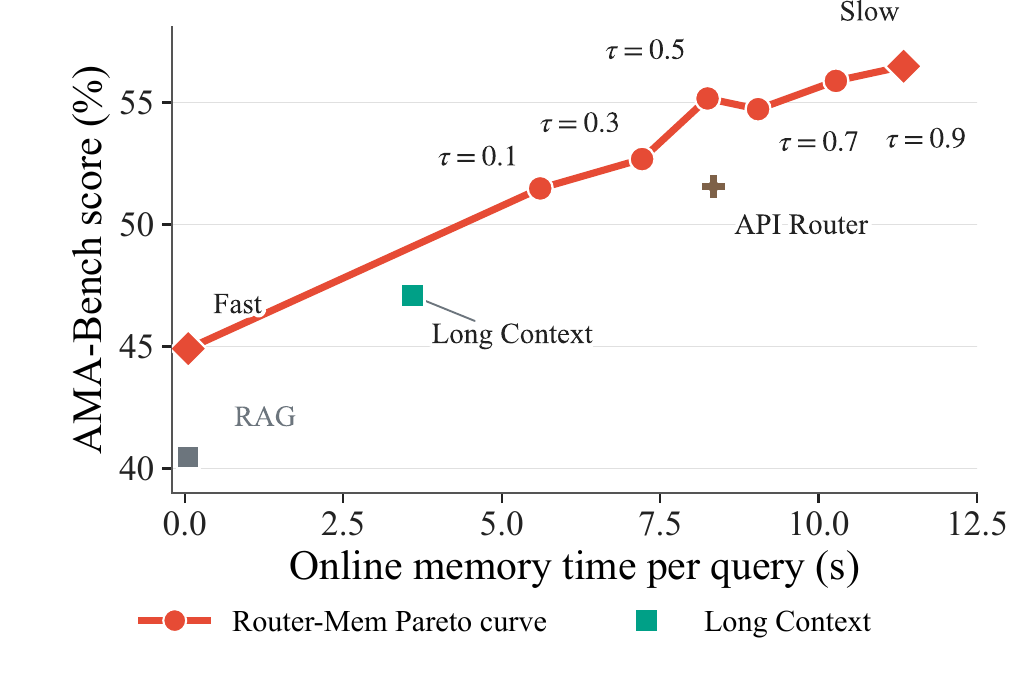}
    \caption{Quality--latency under different routing thresholds.}
    \label{fig:pareto}
\end{figure}

\subsection{Main Results.}
We compare Router-Mem with representative memory mechanisms from three categories. 
RAG serves as a lightweight retrieval baseline that directly retrieves relevant contexts without persistent memory modeling. 
LightMem and Mem0 represent compact memory systems that extract and organize historical information into efficient memory representations. 
Long Context provides a full-context baseline that directly exposes the entire history to the language model. 
We further compare against two execution endpoints: Fast Prefix Only, which always terminates after the shared retrieval stage, and Full Completion (Slow), which always performs complete memory reconstruction. 
These endpoints characterize the lower-cost and upper-performance boundaries of progressive memory execution.

\paragraph{Results on AMA-Bench.}
Table~\ref{tab:ama_main} shows that Router-Mem consistently achieves a favorable quality--latency trade-off between the fast and slow endpoints. 
The Fast Prefix Only baseline achieves the lowest memory latency ($0.054$s) but suffers from incomplete evidence coverage, obtaining 44.91\% score. 
In contrast, Full Completion reaches the highest score of 56.49\% with substantially larger latency. 
By adaptively routing queries, Router-Mem approaches the performance of full completion while avoiding unnecessary deep memory processing. 
For example, Router-Mem ($\tau=0.5$) achieves 55.17\% score with only 8.25s memory latency, reducing the cost by 27.3\% compared with full completion. 
Different thresholds provide controllable operating points, demonstrating that the learned router effectively balances answer quality and execution efficiency.

\paragraph{Results on BEAM.}
Results on BEAM further demonstrate the scalability of Router-Mem under extremely long conversational histories. 
The Fast Prefix Only baseline obtains only 30.02\% score, while Full Completion improves performance to 43.26\% at the cost of 14.61s memory latency. 
Router-Mem bridges this gap by allocating deeper reasoning only when necessary. 
With $\tau=0.7$, it achieves 39.81\% score while reducing memory latency to 12.51s, and with $\tau=0.9$, it approaches the full completion performance (42.09\%) with lower inference cost. 
Across thresholds, Router-Mem forms a smooth quality--cost frontier, validating that evidence-conditioned routing enables adaptive computation allocation for long-term agent memory.

\begin{table}[t]
\centering
\caption{
Router training comparison on AMA-Bench and BEAM.
ET denotes the early-termination rate.
}
\label{tab:router-ablation-combined}

\begingroup
\scriptsize
\renewcommand{\arraystretch}{0.92}
\setlength{\tabcolsep}{6.2pt}
\setlength{\aboverulesep}{1.2pt}
\setlength{\belowrulesep}{1.2pt}

\begin{tabular}{clrrr rrr}
\toprule
& &
\multicolumn{3}{c}{AMA-Bench}
& \multicolumn{3}{c}{BEAM} \\
\cmidrule(lr){3-5}
\cmidrule(lr){6-8}
$\tau$
& Router
& Score 
& ET 
& $T_{\mathrm{e2e}}$ 
& Score 
& ET 
& $T_{\mathrm{e2e}}$  \\
&
& (\%)
& (\%)
& (s)
& (\%)
& (\%)
& (s) \\
\midrule

0.3
& Base
& 55.63
& 8.0
& 13.68
& 29.71
& 90.5
& 4.15 \\
& Trained
& 52.68
& 38.5
& 10.53
& 36.53
& 39.4
& 11.72 \\
\midrule

0.5
& Base
& 55.52
& 4.9
& 14.21
& 29.62
& 86.5
& 5.15 \\
& Trained
& 55.17
& 30.0
& 11.57
& 38.77
& 26.6
& 14.11 \\
\midrule

0.7
& Base
& 55.79
& 2.2
& 14.39
& 31.78
& 80.5
& 5.67 \\
& Trained
& 54.73
& 22.1
& 12.26
& 39.81
& 16.7
& 16.18 \\

\bottomrule
\end{tabular}
\endgroup
\end{table}

\begin{table}[t]
\centering
\caption{
Effect of embedding training on AMA-Bench.
}
\label{tab:embedding-ablation}
\small
\setlength{\tabcolsep}{4pt}
\resizebox{\columnwidth}{!}{
\begin{tabular}{llrrrr}
\toprule
Setting
& Embedding
& Score (\%) 
& Fast hit (\%) 
& $T_{\mathrm{mem}}$ (s) 
& $T_{\mathrm{e2e}}$ (s)  \\
\midrule
Fast-only
& Base
& 42.00
& 100.0
& 0.055
& 3.41 \\
Fast-only
& Trained
& 44.91
& 100.0
& 0.054
& 3.70 \\
\midrule
Router ($\tau=0.5$)
& Base
& 54.79
& 21.8
& 9.17
& 12.16 \\
Router ($\tau=0.5$)
& Trained
& 55.17
& 30.0
& 8.25
& 11.57 \\
\midrule
Slow-only
& Base
& 55.73
& 0.0
& 11.29
& 14.32 \\
Slow-only
& Trained
& 56.49
& 0.0
& 11.34
& 14.47 \\
\bottomrule
\end{tabular}
}
\end{table}
\subsection{Ablation Study}
% \paragraph{Memory Performance and Lantancy Trade-off} Figure~\ref{fig:pareto} illustrates how the routing threshold controls the
% quality--latency trade-off on AMA-Bench. Increasing $\tau$ makes the router
% more conservative and sends more queries to the deeper memory-processing
% path. This generally improves answer quality at the cost of additional
% online memory time. Among the resulting operating points, $\tau=0.5$
% provides a particularly favorable Pareto-efficient trade-off. It achieves
% a score of 55.17\%, only 1.32 points below the Full Completion endpoint
% (56.49\%), while reducing $T_{\mathrm{mem}}$ from 11.341s to 8.248s
% and $T_{\mathrm{e2e}}$ from 14.47s to 11.57s. These correspond to
% 27.3\% and 20.0\% latency reductions, respectively. Thus, $\tau=0.5$
% retains most of the accuracy benefit of full memory reconstruction while
% avoiding a substantial fraction of its computational overhead.

\paragraph{Memory Performance and Latency Trade-off.}
Figure~\ref{fig:pareto} illustrates how the routing threshold controls the
quality--latency trade-off on AMA-Bench. Increasing $\tau$ makes the router
more conservative and sends more queries to the deeper memory-processing
path. This generally improves answer quality at the cost of additional
online memory time. Among the resulting operating points, $\tau=0.5$
provides a particularly favorable Pareto-efficient trade-off. At a
comparable online memory time, this operating point also achieves a
substantially higher AMA-Bench score than the large model API Router baseline. It
achieves a score of 55.17\%, only 1.32 points below the Full Completion
endpoint (56.49\%), while reducing $T_{\mathrm{mem}}$ from 11.341s to
8.248s and $T_{\mathrm{e2e}}$ from 14.47s to 11.57s. These correspond to
27.3\% and 20.0\% latency reductions, respectively. Thus, $\tau=0.5$
retains most of the accuracy benefit of full memory reconstruction while
avoiding a substantial fraction of its computational overhead.

\paragraph{Effect of Router Training.}
As shown in Table~\ref{tab:router-ablation-combined}, router training primarily calibrates the fast--slow decision boundary rather than uniformly favoring one execution path. On AMA-Bench, the base router is strongly biased toward the slow path, with only 4.9\% early termination at $\tau=0.5$. Training raises this rate to 30.0\% and reduces $T_{\mathrm{e2e}}$ from 14.21s to 11.57s, while preserving a comparable score (55.52\% vs.\ 55.17\%). In contrast, the base router on BEAM is over-confident in the fast path, terminating 86.5\% of queries early and achieving only 29.62\%. Training corrects this bias, reducing early termination to 26.6\% and improving the score to 38.77\%. Moreover, the trained router responds consistently to changes in $\tau$, enabling flexible control over computation. Without training, the decision boundary collapses toward opposite fixed endpoints across the two benchmarks, making threshold adjustment largely ineffective.

\paragraph{Effect of Embedding Training.}
We study the effect of training the embedding model on AMA-Bench.
As shown in Table~\ref{tab:embedding-ablation}, the trained embedding
improves the Fast-only score from 42.00 to 44.91 without changing the memory latency. Under the practical Router-Mem setting at
$\tau=0.5$, it increases the score from 54.79 to 55.17 and the fast-hit
rate from 21.8\% to 30.0\%, while reducing $T_{\mathrm{mem}}$ from
9.17s to 8.25s and $T_{\mathrm{e2e}}$ from 12.16s to 11.57s.
The Slow-only endpoint also gains 0.76 points with nearly unchanged
runtime. These results indicate that embedding training primarily
improves retrieved evidence quality and routing efficiency rather than
increasing computation.

\begin{table}[t]
\centering
\caption{Effect of jointly training the router and retriever on AMA-Bench.}
\label{tab:joint-training}

\begingroup
\scriptsize
\renewcommand{\arraystretch}{0.90}
\setlength{\tabcolsep}{2.0pt}
\setlength{\aboverulesep}{1.5pt}
\setlength{\belowrulesep}{1.5pt}

\begin{tabular}{clrrrr}
\toprule
$\tau$
& Training
& Score (\%) 
& Early Term. (\%) 
& $T_{\mathrm{mem}}$ (s) 
& $T_{\mathrm{e2e}}$ (s)  \\
\midrule

0.3
& None
& 54.48
& 9.4
& 10.352
& 13.453 \\
& Joint
& 52.68
& 38.5
& 7.215
& 10.534 \\
\midrule

0.5
& None
& 53.96
& 6.1
& 10.842
& 13.987 \\
& Joint
& 55.17
& 30.0
& 8.248
& 11.570 \\
\midrule

0.7
& None
& 56.25
& 3.5
& 11.073
& 14.219 \\
& Joint
& 54.73
& 22.1
& 9.045
& 12.260 \\

\bottomrule
\end{tabular}
\endgroup
\end{table}

\paragraph{Effect of Joint Training.}
We also evaluate the joint effect of training both the router and retriever.
As shown in Table~\ref{tab:joint-training}, without training, the system
behaves similarly to the Slow-only endpoint: early termination remains
between 3.5\% and 9.4\%, and varying $\tau$ provides limited control over
computation. Joint training increases this range to 22.1\%--38.5\% and
consistently reduces both memory and end-to-end latency while preserving
comparable answer quality. At $\tau=0.5$, it improves the score from
53.96\% to 55.17\%, raises early termination from 6.1\% to 30.0\%, and
reduces $T_{\mathrm{mem}}$ from 10.842s to 8.248s. This broader operating
range indicates that the learned components jointly calibrate retrieval
quality and termination confidence across thresholds. These results show that joint training more reliably identifies safely
answerable queries and makes $\tau$ an effective control knob for the
quality--latency trade-off.

\section{Conclusion}

% In this paper, we propose Router-Mem to address the challenge of efficiently accessing long-horizon memories in LLM agents, where existing methods often trade off between low-cost retrieval and expensive deep memory reasoning. Router-Mem is an evidence-conditioned progressive memory execution framework that learns when the current retrieved context is sufficient to terminate further computation. It first performs a shared retrieval prefix and employs a lightweight sufficiency router to make an early-termination decision. When evidence is insufficient, it reuses retrieval anchors to guide memory expansion and deeper evidence aggregation. Experiments on AMA-Bench and BEAM demonstrate that Router-Mem achieves strong answer quality while substantially reducing inference cost. We believe our approach provides a practical direction toward adaptive memory systems that allocate reasoning resources according to the actual evidence requirements of each query.
In this paper, we propose Router-Mem to address the challenge of efficiently accessing long-horizon memories in LLM agents, where existing methods often trade off between low-cost retrieval and expensive deep memory reasoning. Router-Mem is an evidence-conditioned progressive memory execution framework that learns when the current retrieved context is sufficient to terminate further computation. It first performs a shared retrieval prefix and employs a lightweight sufficiency router to make an early-termination decision. In particular, the shared prefix couples the fast and slow paths, allowing the system to preserve retrieved evidence and avoid restarting memory search when deeper processing is required. When evidence is insufficient, it reuses retrieval anchors to guide memory expansion and deeper evidence aggregation. Experiments on AMA-Bench and BEAM demonstrate that Router-Mem achieves strong answer quality while substantially reducing inference cost. We believe our approach provides a practical direction toward adaptive memory systems that allocate reasoning resources according to the actual evidence requirements of each query.

\bibliography{aaai2027}

@inproceedings{lewis2020rag,
  title     = {Retrieval-Augmented Generation for Knowledge-Intensive NLP Tasks},
  author    = {Lewis, Patrick and Perez, Ethan and Piktus, Aleksandra and
               Petroni, Fabio and Karpukhin, Vladimir and Goyal, Naman and
               K{\"u}ttler, Heinrich and Lewis, Mike and Yih, Wen-tau and
               Rockt{\"a}schel, Tim and Riedel, Sebastian and Kiela, Douwe},
  booktitle = {Advances in Neural Information Processing Systems},
  year      = {2020}
}

@inproceedings{gutierrez2024hipporag,
  title     = {{HippoRAG}: Neurobiologically Inspired Long-Term Memory for
               Large Language Models},
  author    = {Guti{\'e}rrez, Bernal Jim{\'e}nez and Shu, Yiheng and Gu, Yu and
               Yasunaga, Michihiro and Su, Yu},
  booktitle = {Advances in Neural Information Processing Systems},
  year      = {2024}
}

@article{edge2024graphrag,
  title   = {From Local to Global: A Graph {RAG} Approach to
             Query-Focused Summarization},
  author  = {Edge, Darren and Trinh, Ha and Cheng, Newman and Bradley, Joshua and
             Chao, Alex and Mody, Apurva and Truitt, Steven and
             Metropolitansky, Dasha and Ness, Robert Osazuwa and Larson, Jonathan},
  journal = {arXiv preprint arXiv:2404.16130},
  year    = {2024}
}

@article{guo2024lightrag,
  title   = {{LightRAG}: Simple and Fast Retrieval-Augmented Generation},
  author  = {Guo, Zirui and Xia, Lianghao and Yu, Yanhua and Ao, Tu and
             Huang, Chao},
  journal = {arXiv preprint arXiv:2410.05779},
  year    = {2024}
}

@article{chhikara2025mem0,
  title   = {{Mem0}: Building Production-Ready {AI} Agents with Scalable
             Long-Term Memory},
  author  = {Chhikara, Prateek and Khant, Dev and Aryan, Saket and
             Singh, Taranjeet and Yadav, Deshraj},
  journal = {arXiv preprint arXiv:2504.19413},
  year    = {2025}
}

@inproceedings{xu2025amem,
  title     = {{A-MEM}: Agentic Memory for {LLM} Agents},
  author    = {Xu, Wujiang and Liang, Zujie and Mei, Kai and Gao, Hang and
               Tan, Juntao and Zhang, Yongfeng},
  booktitle = {Advances in Neural Information Processing Systems},
  year      = {2025}
}

@article{zhang2025gmemory,
  title   = {{G-Memory}: Tracing Hierarchical Memory for Multi-Agent Systems},
  author  = {Zhang, Guibin and Fu, Muxin and Wan, Guancheng and Yu, Miao and
             Wang, Kun and Yan, Shuicheng},
  journal = {arXiv preprint arXiv:2506.07398},
  year    = {2025}
}

@inproceedings{ouyang2026reasoningbank,
  title     = {{ReasoningBank}: Scaling Agent Self-Evolving with Reasoning Memory},
  author    = {Ouyang, Siru and Yan, Jun and Hsu, I-Hung and Chen, Yanfei and
               Jiang, Ke and Wang, Zifeng and Han, Rujun and Le, Long T. and
               Daruki, Samira and Tang, Xiangru and Tirumalashetty, Vishy and
               Lee, George and Rofouei, Mahsan and Lin, Hangfei and Han, Jiawei and
               Lee, Chen-Yu and Pfister, Tomas},
  booktitle = {International Conference on Learning Representations},
  year      = {2026}
}

@article{yan2025gam,
  title   = {General Agentic Memory Via Deep Research},
  author  = {Yan, B. Y. and Li, Chaofan and Qian, Hongjin and Lu, Shuqi and
             Liu, Zheng},
  journal = {arXiv preprint arXiv:2511.18423},
  year    = {2025}
}

@misc{wang2026ememmultiagentbasedepisodic,
      title={E-mem: Multi-agent based Episodic Context Reconstruction for LLM Agent Memory}, 
      author={Kaixiang Wang and Yidan Lin and Jiong Lou and Zhaojiacheng Zhou and Bunyod Suvonov and Jie Li},
      year={2026},
      eprint={2601.21714},
      archivePrefix={arXiv},
      primaryClass={cs.AI},
      url={https://arxiv.org/abs/2601.21714}, 
}

@inproceedings{yao2023react,
  title     = {{ReAct}: Synergizing Reasoning and Acting in Language Models},
  author    = {Yao, Shunyu and Zhao, Jeffrey and Yu, Dian and Du, Nan and
               Shafran, Izhak and Narasimhan, Karthik and Cao, Yuan},
  booktitle = {International Conference on Learning Representations},
  year      = {2023}
}

@inproceedings{schick2023toolformer,
  title     = {Toolformer: Language Models Can Teach Themselves to Use Tools},
  author    = {Schick, Timo and Dwivedi-Yu, Jane and Dess{\`i}, Roberto and
               Raileanu, Roberta and Lomeli, Maria and Zettlemoyer, Luke and
               Cancedda, Nicola and Scialom, Thomas},
  booktitle = {Advances in Neural Information Processing Systems},
  year      = {2023}
}

@inproceedings{shinn2023reflexion,
  title     = {Reflexion: Language Agents with Verbal Reinforcement Learning},
  author    = {Shinn, Noah and Cassano, Federico and Berman, Edward and
               Gopinath, Ashwin and Narasimhan, Karthik and Yao, Shunyu},
  booktitle = {Advances in Neural Information Processing Systems},
  year      = {2023}
}

@article{wang2024voyager,
  title   = {Voyager: An Open-Ended Embodied Agent with Large Language Models},
  author  = {Wang, Guanzhi and Xie, Yuqi and Jiang, Yunfan and
             Mandlekar, Ajay and Xiao, Chaowei and Zhu, Yuke and
             Fan, Linxi and Anandkumar, Anima},
  journal = {Transactions on Machine Learning Research},
  year    = {2024}
}

@inproceedings{zhou2024webarena,
  title     = {{WebArena}: A Realistic Web Environment for Building
               Autonomous Agents},
  author    = {Zhou, Shuyan and Xu, Frank F. and Zhu, Hao and Zhou, Xuhui and
               Lo, Robert and Sridhar, Abishek and Cheng, Xianyi and Ou, Tianyue
               and Bisk, Yonatan and Fried, Daniel and Alon, Uri and
               Neubig, Graham},
  booktitle = {International Conference on Learning Representations},
  year      = {2024}
}

@inproceedings{liu2024agentbench,
  title     = {{AgentBench}: Evaluating {LLM}s as Agents},
  author    = {Liu, Xiao and Yu, Hao and Zhang, Hanchen and Xu, Yifan and
               Lei, Xuanyu and Lai, Hanyu and Gu, Yu and Ding, Hangliang and
               Men, Kaiwen and Yang, Kejuan and Zhang, Shudan and Deng, Xiang and
               Zeng, Aohan and Du, Zhengxiao and Zhang, Chenhui and Shen, Sheng
               and Zhang, Tianjun and Su, Yu and Sun, Huan and Huang, Minlie and
               Dong, Yuxiao and Tang, Jie},
  booktitle = {International Conference on Learning Representations},
  year      = {2024}
}

@inproceedings{yang2024sweagent,
  title     = {{SWE-agent}: Agent-Computer Interfaces Enable Automated
               Software Engineering},
  author    = {Yang, John and Jimenez, Carlos E. and Wettig, Alexander and
               Lieret, Kilian and Yao, Shunyu and Narasimhan, Karthik and
               Press, Ofir},
  booktitle = {Advances in Neural Information Processing Systems},
  year      = {2024}
}

@inproceedings{dai2019transformerxl,
  title     = {Transformer-{XL}: Attentive Language Models beyond a
               Fixed-Length Context},
  author    = {Dai, Zihang and Yang, Zhilin and Yang, Yiming and
               Carbonell, Jaime and Le, Quoc and Salakhutdinov, Ruslan},
  booktitle = {Proceedings of the 57th Annual Meeting of the
               Association for Computational Linguistics},
  pages     = {2978--2988},
  year      = {2019}
}

@article{beltagy2020longformer,
  title   = {Longformer: The Long-Document Transformer},
  author  = {Beltagy, Iz and Peters, Matthew E. and Cohan, Arman},
  journal = {arXiv preprint arXiv:2004.05150},
  year    = {2020}
}

@inproceedings{zaheer2020bigbird,
  title     = {Big Bird: Transformers for Longer Sequences},
  author    = {Zaheer, Manzil and Guruganesh, Guru and Dubey, Avinava and
               Ainslie, Joshua and Alberti, Chris and Onta{\~n}{\'o}n, Santiago
               and Pham, Philip and Ravula, Anirudh and Wang, Qifan and
               Yang, Li and Ahmed, Amr},
  booktitle = {Advances in Neural Information Processing Systems},
  year      = {2020}
}

@inproceedings{dao2022flashattention,
  title     = {{FlashAttention}: Fast and Memory-Efficient Exact Attention
               with {IO}-Awareness},
  author    = {Dao, Tri and Fu, Daniel Y. and Ermon, Stefano and
               Rudra, Atri and R{\'e}, Christopher},
  booktitle = {Advances in Neural Information Processing Systems},
  year      = {2022}
}

@article{liu2024lost,
  title   = {Lost in the Middle: How Language Models Use Long Contexts},
  author  = {Liu, Nelson F. and Lin, Kevin and Hewitt, John and
             Paranjape, Ashwin and Bevilacqua, Michele and
             Petroni, Fabio and Liang, Percy},
  journal = {Transactions of the Association for Computational Linguistics},
  volume  = {12},
  pages   = {157--173},
  year    = {2024}
}

@inproceedings{bai2024longbench,
  title     = {{LongBench}: A Bilingual, Multitask Benchmark for
               Long Context Understanding},
  author    = {Bai, Yushi and Lv, Xin and Zhang, Jiajie and Lyu, Hongchang and
               Tang, Jiankai and Huang, Zhidian and Du, Zhengxiao and
               Liu, Xiao and Zeng, Aohan and Hou, Lei and Dong, Yuxiao and
               Tang, Jie and Li, Juanzi},
  booktitle = {Proceedings of the 62nd Annual Meeting of the
               Association for Computational Linguistics},
  pages     = {3119--3137},
  year      = {2024}
}

@inproceedings{hsieh2024ruler,
  title     = {{RULER}: What's the Real Context Size of Your
               Long-Context Language Models?},
  author    = {Hsieh, Cheng-Ping and Sun, Simeng and Kriman, Samuel and
               Acharya, Shantanu and Rekesh, Dima and Jia, Fei and
               Zhang, Yang and Ginsburg, Boris},
  booktitle = {Conference on Language Modeling},
  year      = {2024}
}

@inproceedings{zhang2024infinitebench,
  title     = {$\infty$Bench: Extending Long Context Evaluation
               Beyond 100K Tokens},
  author    = {Zhang, Xinrong and Chen, Yingfa and Hu, Shengding and
               Xu, Zihang and Chen, Junhao and Hao, Moo Khai and
               Han, Xu and Thai, Zhen Leng and Wang, Shuo and
               Liu, Zhiyuan and Sun, Maosong},
  booktitle = {Proceedings of the 62nd Annual Meeting of the
               Association for Computational Linguistics},
  year      = {2024}
}

@article{zhang2025survey,
  title={A survey on the memory mechanism of large language model-based agents},
  author={Zhang, Zeyu and Dai, Quanyu and Bo, Xiaohe and Ma, Chen and Li, Rui and Chen, Xu and Zhu, Jieming and Dong, Zhenhua and Wen, Ji-Rong},
  journal={ACM Transactions on Information Systems},
  volume={43},
  number={6},
  pages={1--47},
  year={2025},
  publisher={ACM New York, NY}
}

@article{hu2025memory,
  title={Memory in the age of ai agents},
  author={Hu, Yuyang and Liu, Shichun and Yue, Yanwei and Zhang, Guibin and Liu, Boyang and Zhu, Fangyi and Lin, Jiahang and Guo, Honglin and Dou, Shihan and Xi, Zhiheng and others},
  journal={arXiv preprint arXiv:2512.13564},
  year={2025}
}

@article{li2024survey,
  title={A survey on LLM-based multi-agent systems: workflow, infrastructure, and challenges},
  author={Li, Xinyi and Wang, Sai and Zeng, Siqi and Wu, Yu and Yang, Yi},
  journal={Vicinagearth},
  volume={1},
  number={1},
  pages={9},
  year={2024},
  publisher={Springer}
}

@article{guo2024large,
  title={Large language model based multi-agents: A survey of progress and challenges},
  author={Guo, Taicheng and Chen, Xiuying and Wang, Yaqi and Chang, Ruidi and Pei, Shichao and Chawla, Nitesh V and Wiest, Olaf and Zhang, Xiangliang},
  journal={arXiv preprint arXiv:2402.01680},
  year={2024}
}

@article{du2026memory,
  title={Memory for autonomous llm agents: Mechanisms, evaluation, and emerging frontiers},
  author={Du, Pengfei},
  journal={arXiv preprint arXiv:2603.07670},
  year={2026}
}

@inproceedings{jeong2024adaptive,
  title     = {Adaptive-{RAG}: Learning to Adapt Retrieval-Augmented
               Large Language Models through Question Complexity},
  author    = {Jeong, Soyeong and Baek, Jinheon and Cho, Sukmin and
               Hwang, Sung Ju and Park, Jong},
  booktitle = {Proceedings of the 2024 Conference of the North American
               Chapter of the Association for Computational Linguistics:
               Human Language Technologies},
  pages     = {7036--7050},
  publisher = {Association for Computational Linguistics},
  year      = {2024}
}

@article{park2025stoprag,
  title   = {{Stop-RAG}: Value-Based Retrieval Control for Iterative {RAG}},
  author  = {Park, Jaewan and Cho, Solbee and Lee, Jay-Yoon},
  journal = {arXiv preprint arXiv:2510.14337},
  year    = {2025}
}

@article{li2026raser,
  title   = {{RASER}: Recoverability-Aware Selective Escalation Router
             for Multi-Hop Question Answering},
  author  = {Li, Yuyang and Yan, Zihe and K{\"a}fer, Tobias},
  journal = {arXiv preprint arXiv:2606.02488},
  year    = {2026}
}

@inproceedings{zhao2026amabench,
  title     = {{AMA-Bench}: Evaluating Long-Horizon Memory for
               Agentic Applications},
  author    = {Zhao, Yujie and Yuan, Boqin and Huang, Junbo and
               Yuan, Haocheng and Yu, Zhongming and Xu, Haozhou and
               Hu, Lanxiang and Shankarampeta, Abhilash and
               Huang, Zimeng and Ni, Wentao and Tian, Yuandong and
               Zhao, Jishen},
  booktitle = {Proceedings of the 43rd International Conference on
               Machine Learning},
  year      = {2026}
}

@inproceedings{tavakoli2026beam,
  title     = {Beyond a Million Tokens: Benchmarking and Enhancing
               Long-Term Memory in {LLM}s},
  author    = {Tavakoli, Mohammad and Salemi, Alireza and Ye, Carrie and
               Abdalla, Mohamed and Zamani, Hamed and
               Mitchell, J. Ross},
  booktitle = {International Conference on Learning Representations},
  year      = {2026}
}

@misc{fang2026lightmemlightweightefficientmemoryaugmented,
      title={LightMem: Lightweight and Efficient Memory-Augmented Generation}, 
      author={Jizhan Fang and Xinle Deng and Haoming Xu and Ziyan Jiang and Yuqi Tang and Ziwen Xu and Shumin Deng and Yunzhi Yao and Mengru Wang and Shuofei Qiao and Huajun Chen and Ningyu Zhang},
      year={2026},
      eprint={2510.18866},
      archivePrefix={arXiv},
      primaryClass={cs.CL},
      url={https://arxiv.org/abs/2510.18866}, 
}


\end{document}